\documentclass{llncs}

\usepackage{graphicx}
\usepackage{epsfig}
\usepackage[ansinew]{inputenc}
\usepackage{amsmath}
\usepackage{amssymb}
\usepackage{blindtext}
\usepackage{subfig}

\graphicspath{{./figures/}}

\begin{document}
\pagestyle{empty}
\mainmatter

\title{Optimization Networks for \\ Integrated Machine Learning}

\author{
	Michael 	 Kommenda\textsuperscript{1,2} \and 
	Johannes	 Karder\textsuperscript{1,2}  \and 
	Andreas		 Beham\textsuperscript{1,2} \and \\
	Bogdan		 Burlacu\textsuperscript{1,2} \and
	Gabriel		 Kronberger\textsuperscript{1} \and \\
	Stefan		 Wagner\textsuperscript{1} \and 
	Michael		 Affenzeller\textsuperscript{1,2}
}

\institute {
	Heuristic and Evolutionary Algorithms Laboratory\\
	University of Applied Sciences Upper Austria\\
	Softwarepark 11, 4232 Hagenberg, Austria\\
	\email{michael.kommenda@fh-hagenberg.at}
	 \vspace{0.2cm}	\and 
	Institute for Formal Models and Verification\\
	Johannes Kepler University  Linz\\
	Altenbergerstr. 69, 4040 Linz, Austria\\
}

\maketitle

\renewcommand{\thefootnote}{}
\footnotetext{\hspace{-0em}
	The final publication is available at \url{https://link.springer.com/chapter/10.1007/978-3-319-74718-7\_47}. \textcopyright \, Springer International Publishing AG 2018.
}
\renewcommand\thefootnote{\arabic{footnote}}

\begin{abstract}
	Optimization networks are a new methodology for holistically solving interrelated problems that have been developed with combinatorial optimization problems in mind. In this contribution we revisit the core principles of optimization networks and demonstrate their suitability for solving machine learning problems. We use feature selection in combination with linear model creation as a benchmark application and compare the results of optimization networks to ordinary least squares with optional elastic net regularization.
	Based on this example we justify the advantages of optimization networks by adapting the network to solve other machine learning problems. Finally, optimization analysis is presented, where optimal input values of a system have to be found to achieve desired output values. Optimization analysis can be divided into three subproblems: model creation to describe the system, model selection to choose the most appropriate one and parameter optimization to obtain the input values. Therefore, optimization networks are an obvious choice for handling optimization analysis tasks.

\end{abstract}

\section{Introduction}
\label{sec:Introduction}

A general optimization methodology for solving interrelated problems has been recently proposed and analyzed by Beham et al. \cite{beham2015optimization}, where the effects of solving integrated knapsack and traveling salesman problems, such as the knapsack constrained profitable tour problem or the orienteering problem, are studied and compared to Lagrangian decomposition.

Building upon these first results, optimization networks, a methodology for solving interrelated optimization problems, have been developed. Optimization networks consist of several nodes that provide (partial) solutions for individual subproblems, which can be aggregated to a complete solution. Hence,  optimization networks are applicable if the overall optimization problem is decomposable into individual subproblems or several distinct optimization steps are necessary to obtain a solution. Although this limitation seems restricting, in practice most problems can be decomposed into subproblems. An illustrative example for the benefit of optimization networks is the optimization of the operations within a factory. The optimization of production schedules increase the overall performance. However, if corresponding warehouse and logistics operations are taken into account and optimized as well, the overall performance can be further increased~\cite{hauder2016integrated}.

\begin{figure}
	\centering
	\includegraphics[width=0.6\textwidth]{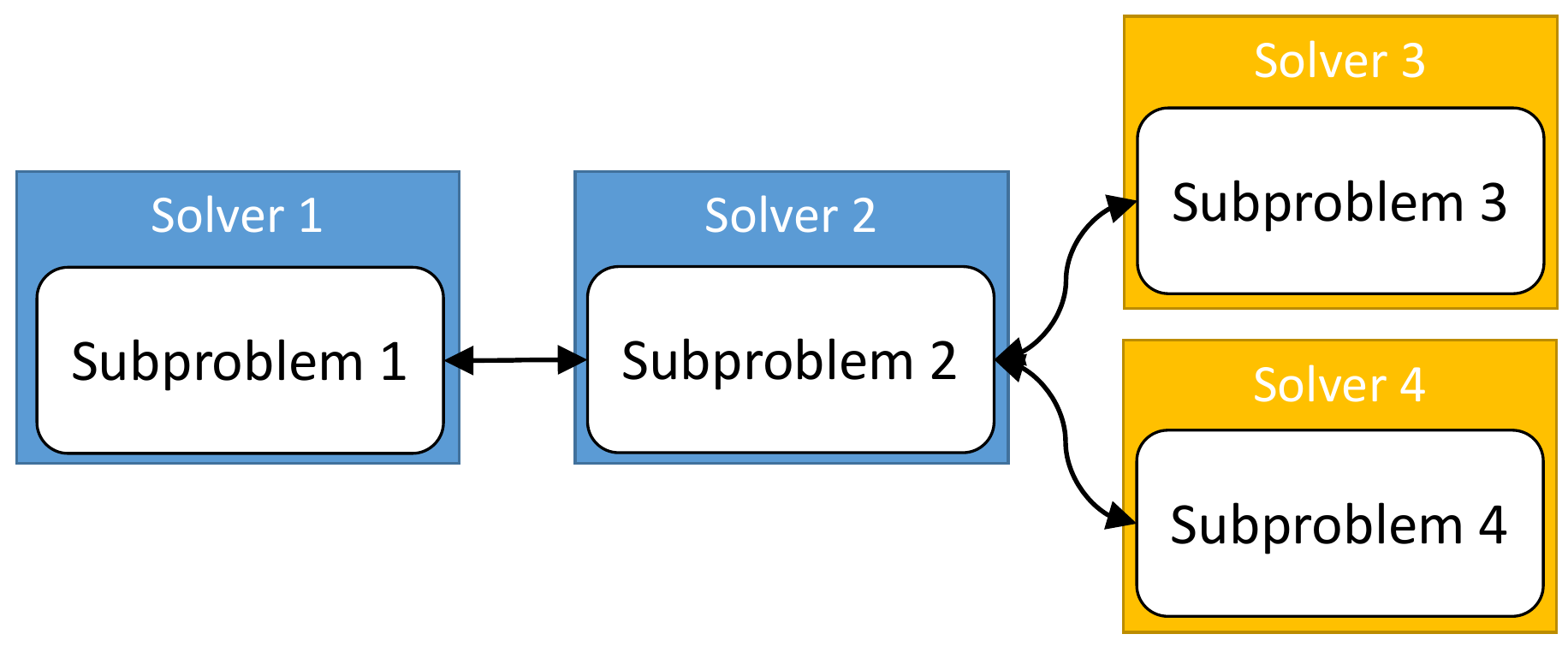}
	\caption{Schematic representation of an optimization network. Each node represents a solver with an according subproblem.}
	\label{fig:ON_schematic}
\end{figure}

The core components of optimization networks are nodes and messages that are sent between nodes through ports \cite{Karder2017}. A node is responsible for performing well defined calculations, for example solving a certain subproblem or transforming data. Ports define the specific interface for communication and which data is sent and received in a message. A restriction is that nodes can only be connected and communicate if the corresponding ports are compatible to each other. In Figure \ref{fig:ON_schematic} a schematic representation of an optimization network is depicted (ports are omitted for clarity). Exchanged messages, indicated by arrows between the nodes, transfer data and steer the execution sequence of the nodes. In this optimization network the nodes termed \emph{Solver~1} and \emph{Solver~2} are executed sequentially, while \emph{Solver~3} and \emph{Solver~4} are executed simultaneously. 

A major benefit of optimization networks is that they enable the reuse of existing nodes and promote the creation of a library of building blocks for algorithm development. To the best of our knowledge, optimization networks have only been applied to combinatorial optimization problems, mostly in the context of production and logistics. In this publication, we demonstrate the suitability of optimization networks for integrated machine learning problems.

\section{Integrated Machine Learning}
\label{sec:IntegratedMachineLearning}

Machine learning enables computers to learn from and make predictions based on data. Whenever multiple, interrelated and dependent machine learning problems are considered in combination we use the term integrated machine learning. Integrated machine learning problems consist of at least two interrelated subproblems, hence are a perfect application area for optimization networks.

Especially when performing supervised machine learning such as regression analysis or classification several steps are necessary to achieve the best possible result. These steps include data preprocessing and cleaning, feature construction and feature selection, model selection and validation, and parameter optimization and are dependent on each other. If the data for creating the models has not been cleaned and preprocessed accordingly, even the best machine learning algorithm will have difficulties to produce accurate and well generalizing prediction models. 

An approach to solve integrated machine learning problems is the automatic construction of tree-based pipelines \cite{olson2016automating}. Those pipelines are represented as trees, where every processing step corresponds to a tree node and genetic programming~\cite{Koza1992} is used for the automatic construction and optimization of those pipelines. This concept is similar to optimization networks due to the reuse and combination of particular building blocks that cooperatively solve the machine learning problem at hand. A difference is that these tree-based pipelines are specific to machine learning, whereas optimization networks are more generic and applicable to diverse optimization problems. However, the versatility of optimization networks comes at a prices, namely that they are created manually by combining already existing building blocks, while machine learning pipelines can be generated automatically.

\section{Feature Selection}
\label{sec:FeatureSelection}

A common supervised machine learning task is regression analysis, where a model describing the relationship between several independent and one dependent variable is built. Another important task is feature subset selection \cite{guyon2003introduction} that determines which of the available independent variables should be used in the model. Both problems are interrelated in the sense that only if an appropriate feature subset is selected, the resulting regression model achieves a good prediction accuracy on training and test data.

An integrated machine learning problem is generating an optimal linear model using the most appropriate features from the data at hand. In total there are $2^{n}$ ($n$ ...  number of features) different linear models possible, ranging from the most simplistic model without any selected feature, which just predicts a constant value, to the most complex one that contains all available features. While this problem can be exhaustively solved for a small number of features, this quickly becomes infeasible due to the exponential growth of possible models.

\begin{figure}[t]
	\centering
	\includegraphics[width=1\textwidth]{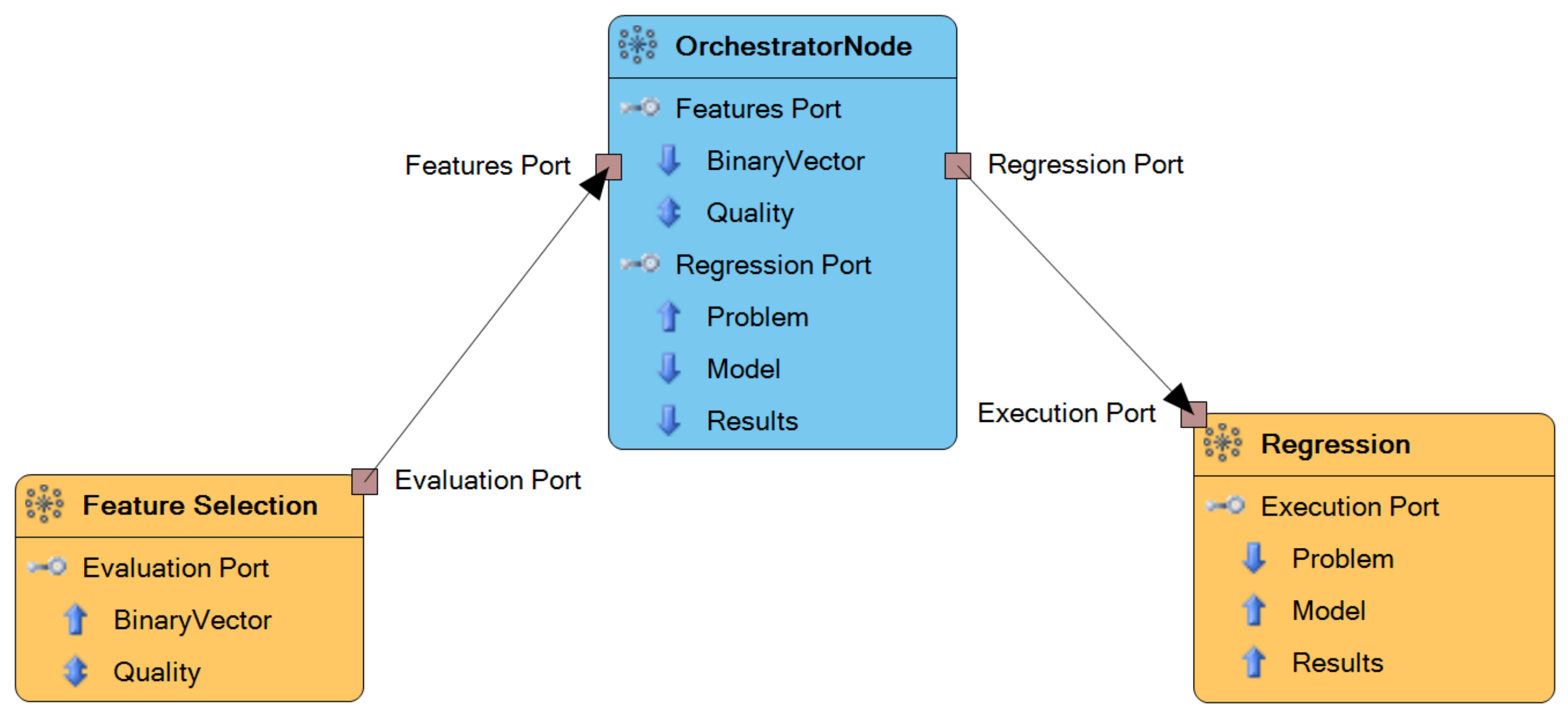}
	\caption{Implementation of an optimization network combining feature selection with regression analysis. When the orchestrator receives a binary vector encoding the selected features, it translates this information to a regression problem that is forwarded to the regression node, which returns the generated model and its quality.}
	\label{fig:ON_FeatureSelection_implementation}
\end{figure}

An optimization network solving the described integrated machine learning problem is shown in Figure \ref{fig:ON_FeatureSelection_implementation}. The feature selection node is only dependent on the number of features in the regression problem and selected features are encoded in a binary vector. To assess the quality of selected features this binary vector is sent to the orchestrator via the connection between the evaluation and features port. The orchestrator parses the binary vector and creates a new regression problem, which only contains selected features, before forwarding it to the regression node. Upon receiving a message with the configured problem, the regression node builds a model and returns it together with its prediction accuracy. This information is passed through the orchestrator to the feature selection node and another combination of selected features can be evaluated. An important aspect is that the feature selection and regression are already existing building blocks and only the orchestrator has to be configured to build the optimization network at hand. 
 
The effectiveness of the proposed optimization network is evaluated on an artificial benchmark problem and compared to standard regression methods creating linear models. Therefore, we created a dataset containing $100$ features $x_i$, where each of the $1000$ observations is sampled from a standard normal distribution $\mathcal{N}(\mu=0,\sigma =1)$. The dependent variable $y$ is calculated as a linear combination of $15$ randomly selected features with uniformly sampled $\mathcal{U}(0,10)$ feature weights $w_i$ and with a normally distributed noise term~$\epsilon$ added that accounts for $20\%$ of the variance of $y$.

A linear model created by ordinary least squares (OLS) \cite{draper1966applied} includes all $100$ features and yields a mean absolute error (MAE) of $10.634$ on a separate test partition. The optimal linear model including only the $15$ necessary inputs has a test MAE of $8.189$. 

Another method for fitting linear models is the minimization of squared errors in combination with elastic net regularization \cite{zou2005regularization}, which balances lasso and ridge regression. A grid search over the parameters $\lambda$ and $p$ is performed to balance the regularization. The best linear model yields a MAE of $8.369$ utilizing $50$ features (only five of the 15 selected features are correctly identified). Elastic net executions with higher regularization yield smaller models, but those including fewer than $20$ features could not identify any feature correctly and therefore have a rather large MAE of approximately $10$.

The optimization network uses an offspring selection genetic algorithm (OSGA) \cite{Affenzeller2009} for generating binary vectors representing the currently selected features and OLS for creating the linear models. The quality of selected features is the prediction error in terms of the mean absolute error on a separate validation partition and a regularization term accounting for the number of selected features. Due to the OSGA being a stochastic metaheuristic, the optimization network has been executed $50$ times. On average $12414$ solutions have been created by the optimization network, which took slightly over one minute for each repetition. All of the repetitions except one identified linear models with $14$ correct and only one missing feature and a MAE of $8.216$ on the test partition. A further investigation revealed that the impact of the missing feature on the model evaluation is lower than the noise level due to a very small feature weight, hence it has not been correctly identified. These results on the described benchmark problem exemplarily demonstrate the benefits and suitability of optimization networks for integrated machine learning problems.

\section{Further Applications}
\label{sec:FurtherApplications}
\begin{figure}
	\centering
	\subfloat[][]{
		\includegraphics[width=0.48\textwidth]{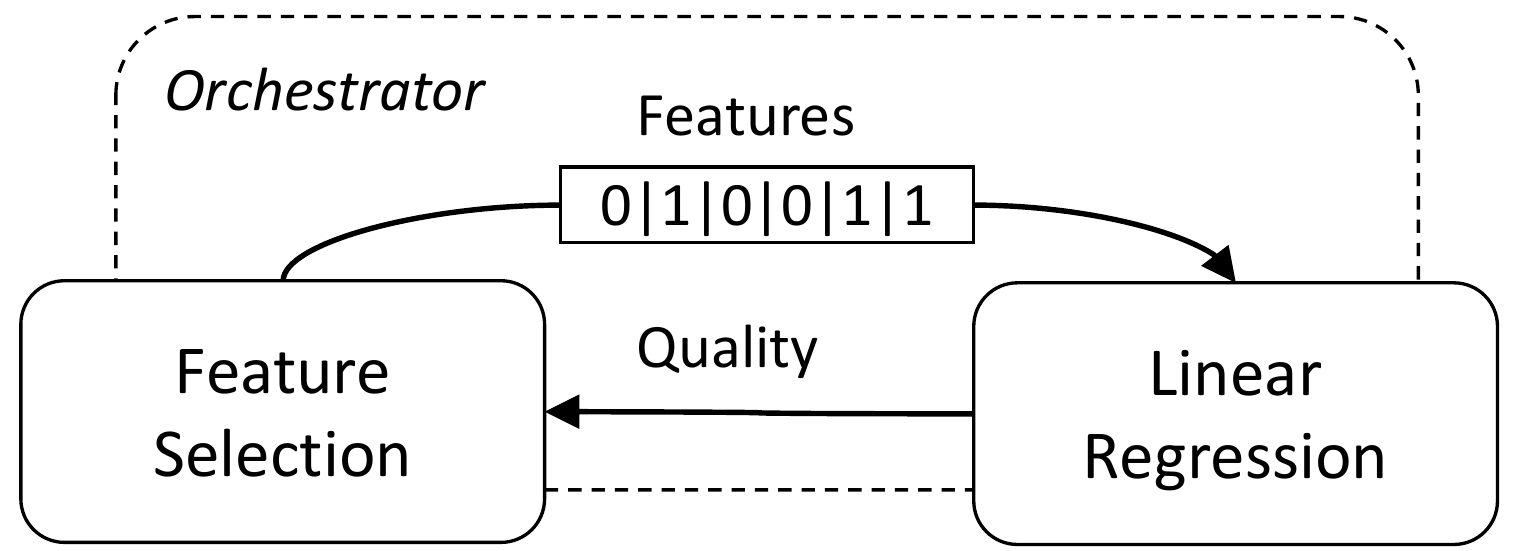}
		\label{fig:ON_FeatureSelection_LinearRegression}
	}
	\subfloat[][]{
		\includegraphics[width=0.48\textwidth]{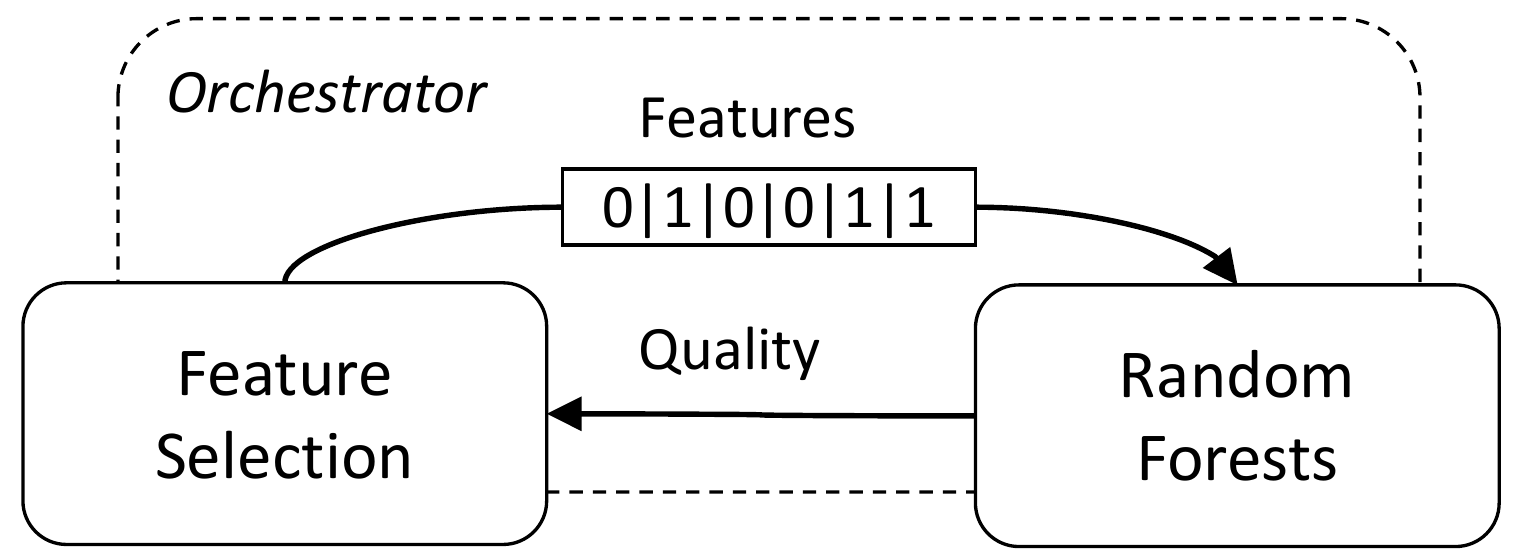}
		\label{fig:ON_FeatureSelection_RandomForests}
	}
	\caption{Optimization networks combining feature selection with either linear regression \protect\subref{fig:ON_FeatureSelection_LinearRegression} or random forest regression \protect\subref{fig:ON_FeatureSelection_RandomForests}.}
	\label{fig:ON:_FeatureSelection_Algorithms}
\end{figure}

In the previous sections, optimization networks and integrated machine learning have been introduced and the performance of optimization networks has been demonstrated using a benchmark problem. Although the obtained results are satisfactory, the full potential of optimization networks for machine learning is only reached in combination with a library of preconfigured building blocks (network nodes). With such a library new optimization networks can be easily created by combining existing nodes and only the orchestrators responsible for controlling the execution sequence and data conversion and transfer have to be implemented anew. A conceptual version of the optimization network for feature selection and linear regression is depicted in Figure \ref{fig:ON_FeatureSelection_LinearRegression}. By exchanging the node for model creation (Figure \ref{fig:ON_FeatureSelection_RandomForests}), the adapted network learns a forest of decision trees \cite{breiman2001random}  instead of linear models.

\begin{figure}[]
	\centering
	\includegraphics[width=1\textwidth]{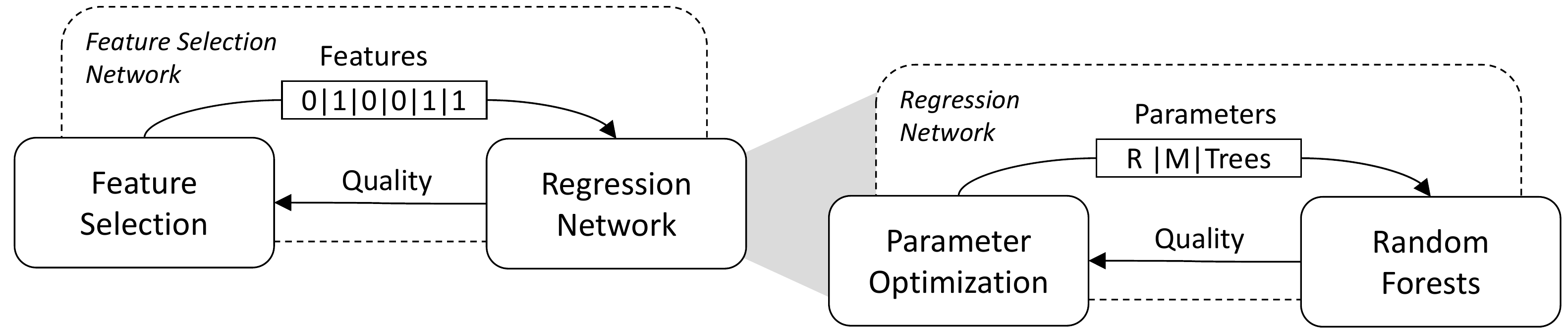}
	\caption{Reuse of optimization networks is facilitated by their inclusion as nodes in other networks. In this example the outer optimization network performs feature selection while the nested network is responsible for parameter optimization of the regression algorithm and model creation.}
	\label{fig:ON_FeatureSelection_GridSearch}
\end{figure}

Additionally to modifying existing optimization networks by exchanging compatible nodes, reuse of components is further enabled by the inclusion of optimization networks as nodes of other networks. As an example, the right optimization network in Figure \ref{fig:ON_FeatureSelection_GridSearch} creates a random forest regression model and simultaneously optimizes the parameters of the algorithm, specifically the ratio of training samples $R$, the ratio of features used for selecting the split $M$, and the number of trees to obtain the most accurate prediction model. Should this network perform feature selection as well, there is no need to adapt it by including an additional feature selection node and the according orchestrator. A better way is to modify the existing feature selection network, similarly to the changes indicated in Figure \ref{fig:ON:_FeatureSelection_Algorithms}, by replacing the linear regression node by the whole regression network. These two applications demonstrate how existing optimization networks can be reused and adapted for solving new problems and therefore highlight the flexibility of optimization networks.

Another application for optimization networks is optimization analysis \cite{o2005introduction}, where values for input features have to be obtained so that predefined output values are achieved. In a previous publication \cite{kommenda2015} the best input parameter combinations of a heat treatment process are obtained so that the quality measures of the final product are optimal within certain limits.
One research question has been which regression models best describe the relation between the input and output values and are therefore most suited for process parameter estimation and optimization analysis. The whole process of regression model creation, model selection and judging the suitability for parameter estimation has been performed manually, but can easily be represented and executed by an optimization network.

\begin{figure}[t]
	\centering
	\includegraphics[width=1\textwidth]{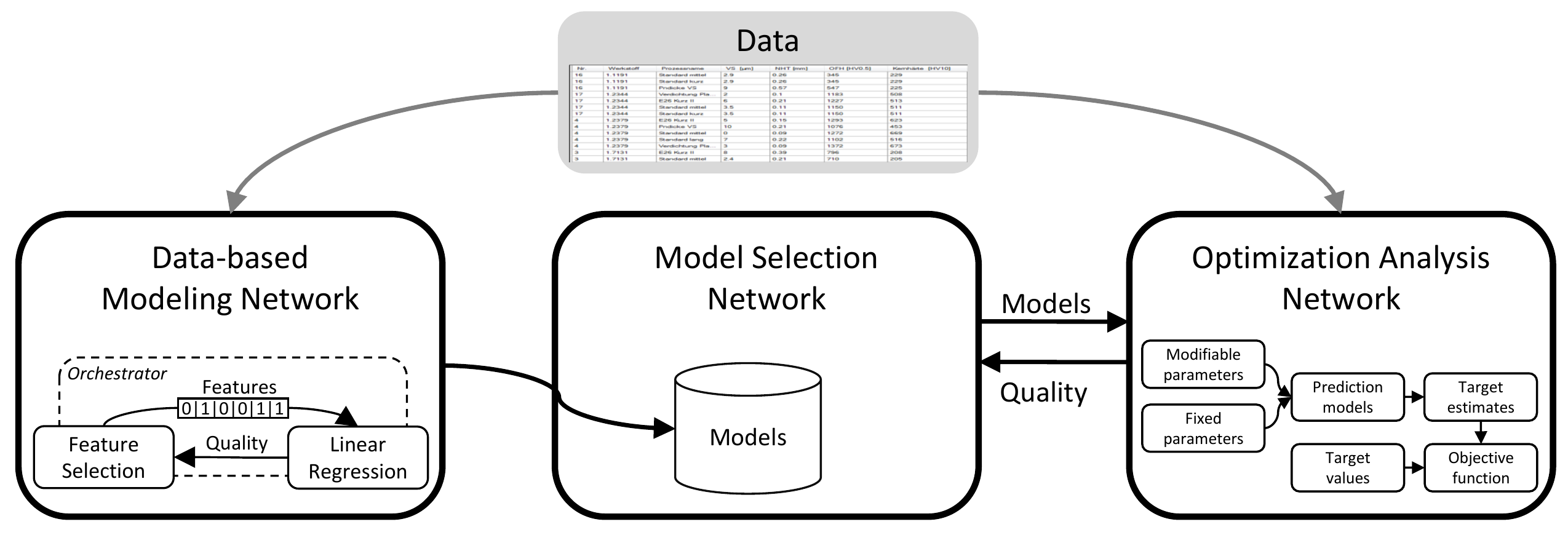}
	\caption{Schematic representation of an optimization network that contains three nested networks that create regression models (left network), select a subset of all generated models (middle network), and evaluate the suitability for optimization analysis (right network).}
	\label{fig:ON_GoalSeeking}
\end{figure}

\newpage
An optimization network combining data-based modeling with optimization analysis is illustrated in Figure \ref{fig:ON_GoalSeeking}. It consists of three nested networks, each tackling a different subproblem. The first network is responsible for data-based modeling and regression model creation. It includes exemplarily the already introduced feature selection and linear regression network, but in practice different machine learning and regression methods are utilized in parallel for creating different prediction models. The second network creates different model subsets, which are forwarded for optimization analysis to the third network. During the whole optimization process the best performing model subset as well as the optimal input features are tracked and accessible to the user for further analysis. With the support of optimization networks the manual effort for performing optimization analysis can be drastically reduced and the combination of different algorithms (regression modeling, subset selection, and parameter optimization) is facilitated.

\section{Conclusion}
\label{sec:Conclusion}
Optimization networks are a new method for solving integrated machine learning problems consisting of several interrelated subproblems. Especially, the reusability and composition of optimization networks from existing building blocks (either configured nodes or whole networks) is a major benefit and eases problem solving. The performance of optimization networks has been demonstrated on a feature selection problem where linear models have to be created. An implementation of optimization networks is available in  HeuristicLab \cite{Wagner2014}, however the number of predefined network nodes is still limited. Therefore, the next step for broadening the scope of optimization networks is to extend and provide a library of building blocks for solving machine learning problems.

\newpage
\section*{Acknowledgments}
\label{subsec:Acknowledgments}
The authors gratefully acknowledge financial support by the Austrian Research Promotion Agency (FFG) and the Government of Upper Austria within the COMET Project \#843532 Heuristic Optimization in Production and Logistics (HOPL).\\

\bibliography{Eurocast2017_Kommenda}

\end{document}